\documentclass[10pt,twocolumn,letterpaper]{article}

\usepackage{cvpr}              
\usepackage{wrapfig} 
\usepackage{multirow}
\usepackage{array}

\definecolor{cvprblue}{rgb}{0.21,0.49,0.74}
\usepackage[pagebackref,breaklinks,colorlinks,allcolors=cvprblue]{hyperref}

\def\confName{CVPR}
\def\confYear{2026}

\title{NESTOR: A Nested MOE-based Neural Operator for Large-Scale PDE Pre-Training}

\author{Dengdi Sun$^{1}$, Xiaoya Zhou$^{1}$, Xiao Wang$^{2}$\thanks{Corresponding Author: Xiao Wang (xiaowang@ahu.edu.cn)}, 
        Hao Si$^{2}$, Wanli Lyu$^{2}$, Jin Tang$^{2}$, Bin Luo$^{2}$ \\ 
${^1}${School of Artificial Intelligence, Anhui University, Hefei, China} \\ 
${^2}${School of Computer Science and Technology, Anhui University, Hefei, China} \\ 
\url{https://github.com/Event-AHU/OpenFusion} 
}

\begin{document}
\maketitle

\begin{abstract}
Neural operators have emerged as an efficient paradigm for solving PDEs, overcoming the limitations of traditional numerical methods and significantly improving computational efficiency. However, due to the diversity and complexity of PDE systems, existing neural operators typically rely on a single network architecture, which limits their capacity to fully capture heterogeneous features and complex system dependencies. This constraint poses a bottleneck for large-scale PDE pre-training based on neural operators. To address these challenges, we propose a large-scale PDE pre-trained neural operator based on a nested Mixture-of-Experts (MoE) framework. In particular, the image-level MoE is designed to capture global dependencies, while the token-level Sub-MoE focuses on local dependencies. Our model can selectively activate the most suitable expert networks for a given input, thereby enhancing generalization and transferability. We conduct large-scale pre-training on twelve PDE datasets from diverse sources and successfully transfer the model to downstream tasks. Extensive experiments demonstrate the effectiveness of our approach. 
\end{abstract}

\section{Introduction}

\begin{figure}
    \centering
    \includegraphics[width=1\linewidth]{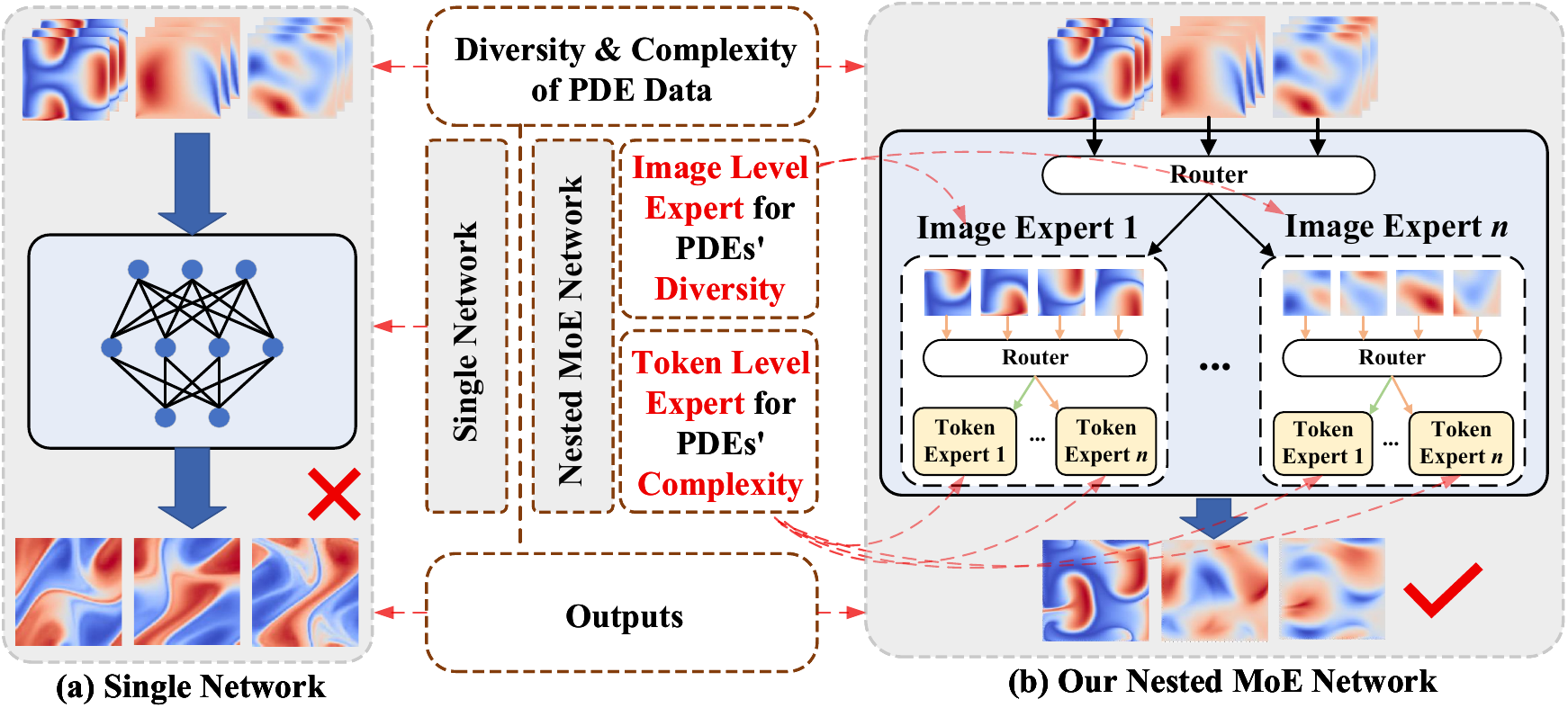}
    \caption{Comparison of two different network architectures. (a) Traditional single-network architecture; (b) our proposed nested MoE architecture, where image-level MoE experts learn global diversity across different PDE types, while token-level Sub-MoE experts capture complex local features within equations.}
    \label{fig:challenge}
\end{figure}

Partial differential equations (PDEs) have broad applications in science and engineering, including physics and fluid mechanics~\cite{karniadakis2021physics}~\cite{debnath2005nonlinear}~\cite{zachmanoglou1986introduction}~\cite{karniadakis2021physics}. Existing studies can be roughly divided into two categories: traditional numerical methods and data-driven methods. Traditional methods, such as FEM~\cite{norrie2014finite} and FDM~\cite{leveque2007finite}, approximate PDE solutions by discretizing the spatial domain, resulting in complex procedures and high computational costs. Neural operators aim to learn infinite-dimensional mappings between function spaces, enabling fast inference while maintaining reasonable accuracy, significantly reducing computational costs, and overcoming the limitations of traditional methods~\cite{li2021neural}\cite{pathak2022fourcastnet}. However, neural operators typically rely on large amounts of training data, which are often obtained through costly experiments and numerical simulations, severely limiting their application in wider scenarios.

Recently, large-scale pre-training~\cite{bengio2012deep} offers a new research paradigm to address this problem. Unlike traditional methods, it involves initially training models on large-scale datasets, enabling them to acquire generalizable knowledge across different PDEs and tasks, thereby establishing a unified modeling framework. For specific downstream tasks, only a small amount of data is required for fine-tuning to obtain highly accurate solutions. This paradigm not only enhances model generalization and effectively mitigates overfitting but also significantly reduces the training cost and time for downstream tasks. Large-scale pre-training has been widely applied in fields such as computer vision and natural language processing~\cite{dosovitskiy2020image}~\cite{devlin2019bert}, where its superior performance has been well validated in practice.

In the field of neural operators~\cite{lu2019deeponet}~\cite{li2020fourier}, research on large-scale pre-training for PDEs has begun to take shape~\cite{hao2024dpot}. However, PDE systems inherently exhibit highly complex spatiotemporal dependencies and significant regional heterogeneity within physical fields. Moreover, different types of PDEs vary substantially in their dynamical mechanisms, boundary conditions, variable dimensions, and numerical distributions. These factors collectively contribute to the diversity and complexity of PDE system data, making unified modeling extremely challenging. Existing large-scale PDE neural operators typically adopt a single network architecture. Although such models can extract general representations shared across different equations, they fall short in capturing the equation-specific characteristics of each PDE type and the localized regional correlations within a single PDE. As illustrated in Fig.~\ref{fig:challenge} , conventional architectures often struggle to simultaneously handle the macroscopic variations across PDEs and the microscopic variations within the same PDE. If the model can incorporate more fine-grained inductive biases into its architecture—thereby learning both the commonalities among different PDEs and the unique properties of each equation, while further identifying local spatial correlations within the physical field—its generalization ability and task adaptability would be significantly enhanced.

In recent years, the Mixture-of-Experts (MoE) framework~\cite{jacobs1991adaptive} has attracted significant attention due to its advantages in increasing model capacity while maintaining computational efficiency. Through the routing mechanism~\cite{jacobs1991adaptive}, MoE can dynamically select the most suitable expert network to provide specialized processing for each input, offering an efficient modeling approach for the pre-training of large-scale PDE neural operators. However, although single-layer MoE models can capture feature differences between equation types, they still face limitations in modeling diversity and complexity within physical fields of the same type of equations.

To address these challenges, we innovatively incorporate the MoE architecture into our model design, constructing a \textbf{NEST}ed MoE-based neural \textbf{O}perato\textbf{R} for large-scale PDE pre-training (\textbf{NESTOR}). Specifically, the image-level MoE adaptively activates the most suitable experts through image-level routing to capture the global features of PDEs. Within each image-level expert, we introduce token-level Sub-MoEs, which selectively activate the most appropriate experts via token-level routing to further capture the complex local correlations within the physical fields. This nested MoE architecture addresses the diversity and complexity of PDEs from both macro and micro perspectives. Through pre-training on large-scale PDE datasets, the architecture can successfully generalize to downstream tasks, providing an efficient modeling and solution framework for complex PDE problems.

The main contributions can be summarized as follows:
\begin{itemize}
\item We propose a novel nested  Mixture-of-Experts architecture that integrates image-level MoE and token-level MoE within a unified framework, enabling cross-level expert collaboration.
\item We designed an image-level routing mechanism that can adaptively select the appropriate expert networks based on the global features of the data, thereby effectively capturing the complex heterogeneous features of different tasks at a global level.
\item Comprehensive validation on large-scale PDE datasets. We apply the proposed framework to large-scale pre-training and downstream tasks across multiple PDE datasets, demonstrating significant advantages in cross-task generalization and transferability.
\end{itemize}

\section{Related Works}
\label{headings}

\subsection{Neural Operators} 
Neural operators are designed to learn mesh-free, function-space-to-function-space infinite-dimensional mappings from inputs to solution functions~\cite{lu2019deeponet}. They effectively overcome the dependence of traditional numerical solvers on mesh discretization, improving computational speed and reducing costs. Moreover, for repeated problems, a neural operator only needs to be trained once, without retraining for each new PDE instance, making it an efficient paradigm for PDE solving. To successfully apply neural operators to PDE problems, researchers have proposed several effective model architectures. For example, DeepONet~\cite{lu2019deeponet} adopts a branch–trunk architecture to realize operator learning. The Fourier Neural Operator (FNO)~\cite{li2020fourier} leverages Fourier transforms to capture non-local dependencies, thus enabling efficient PDE solutions. The Galerkin Transformer~\cite{cao2021choose} integrates self-attention mechanisms with Galerkin projection for operator learning. GNOT~\cite{hao2023gnot} combines graph neural operators with Transformers, achieving efficient modeling on irregular meshes. MPP~\cite{mccabe2023multiple} is a Transformer-based autoregressive pre-training architecture. DPOT~\cite{hao2024dpot} employs autoregressive denoising pre-training combined with Fourier attention to predict a wide range of PDE problems. Poseidon~\cite{herde2024poseidon} integrates neural operators with hybrid attention mechanisms to enable efficient and unified modeling of diverse PDEs. VITO~\cite{ovadia2024vito} integrates Vision Transformers with neural operator principles, enabling vision-based PDE solving and physical field modeling. Unisolver~\cite{zhouunisolver} employs a PDE-conditional Transformer architecture to achieve unified solving across diverse PDEs, advancing toward universal neural PDE solvers. Despite the significant progress made by neural operators, their performance still has room for improvement due to the limitations imposed by the diversity of data and tasks.

\subsection{Mixture of Experts}
The Mixture of Experts (MoE) framework is a method that expands model capacity while avoiding a significant increase in computational cost. Its core idea is to select a subset of experts among multiple expert networks through a gating mechanism ~\cite{jacobs1991adaptive}. With the development of MoE, it has been widely applied in natural language processing, computer vision, and other domains. GShard~\cite{lepikhin2020gshard} was the first to introduce the MoE structure into Transformer models, enabling efficient large-scale distributed training. Switch Transformer~\cite{fedus2022switch} scaled large language model parameters to the trillion level, significantly improving both model capacity and efficiency. V-MoE~\cite{riquelme2021scaling} applied MoE to vision Transformers and demonstrated its potential for enhancing efficiency and performance in tasks such as image recognition. Existing work primarily focuses on homogeneous experts, while research on heterogeneous~\cite{wang2024hmoe} experts is relatively limited. Homogeneous experts refer to all experts using the same network architecture, which offers simplicity in implementation, stable convergence, and ease of load balancing. However, having identical architectures limits expert diversity and, to some extent, constrains the performance of MoE. Heterogeneous expert MoE allows different experts to adopt different network architectures, avoiding redundancy in the features learned by the experts and significantly enhancing the model’s expressive power and efficiency.

\subsection{Pre-training}
Pre-training~\cite{bengio2012deep} refers to the process of training a model on large-scale datasets to learn general knowledge that can be transferred to a variety of downstream tasks. It can significantly reduce the training cost of downstream tasks while improving generalizability. The pre-training paradigm has achieved outstanding success in natural language processing, demonstrating strong cross-task transferability, as exemplified by models such as BERT~\cite{devlin2019bert} and GPT~\cite{radford2018improving}. In computer vision, pre-training has also been widely adopted, with notable examples including the Vision Transformer (ViT)~\cite{dosovitskiy2020image} and CLIP~\cite{radford2021learning}. With the development of large-scale pre-training models, this approach has gradually been introduced into the field of PDE neural operators. Existing explorations include MPP~\cite{mccabe2023multiple}, which proposes a Transformer-based autoregressive pre-training framework capable of learning unified serialized representations across various PDE datasets and allowing cross-task modeling through transfer. DPOT~\cite{hao2024dpot} employs an autoregressive denoising strategy combined with Fourier attention to achieve efficient pre-training across multiple types of PDE problems, demonstrating cross-equation generalization at the operator level. Although these studies have successfully applied pre-training techniques to PDE neural operators, they still exhibit notable limitations in comprehensively capturing PDE systems. Therefore, there remains substantial room for further exploration of large-scale pre-training in the PDE neural operator domain.

\section{Preliminaries}
In this paper, we consider the general form of a parameterized partial differential equation defined on the spatial region $\Omega \subset \mathbb{R}^n$ and the time interval $[0, T]$,
\begin{equation}
\frac{\partial u}{\partial t} - \mathcal{F}\big(u, \nabla u, \nabla^2 u, \dots; \theta \big) = 0,
\end{equation}
\[
\left\{
\begin{array}{l}
u(x,0) = u_0(x), \quad x \in \Omega, \\[0.8ex]
\mathcal{B}[u](x,t) = g(x,t), \quad (x,t) \in \partial \Omega \times (0,T],
\end{array}
\right.
\]
where $u$ is the unknown solution function, representing the state of the system; $\mathcal{F}$ is the PDE spatial derivative operator, which describes the dynamics or evolution law of the system and depends on the current solution u, its spatial derivative, and parameter $\theta$; $\theta$ is the external condition or physical parameter that controls the properties of the equation; u(x,0) is the initial condition; $\mathcal{B}[u](x,t)$ is the boundary condition.

On this basis, we define a solution operator $\mathcal{F}$ and construct the following mapping
\begin{equation}
\mathcal{F}: \quad u_{t+1} = \mathcal{F}_T(u_{t-T+1:t}; \theta),
\end{equation}
where $\theta$ represents the system parameters. The operator $\mathcal{F}$ can take the most recent T frames as input and learn to implicitly infer the details of partial differential equations, the parameters $\theta$, to predict the next frame from the preceding T frames, thereby achieving evolutionary prediction for different system states.

To enhance the model's robustness and generalization, we inject small-scale noise into the input frames. This pretraining strategy has been shown to be effective in DPOT~\cite{hao2024dpot}.

\section{Proposed Method}

\begin{figure*}[t]
    \centering
    \includegraphics[width=0.7\linewidth]{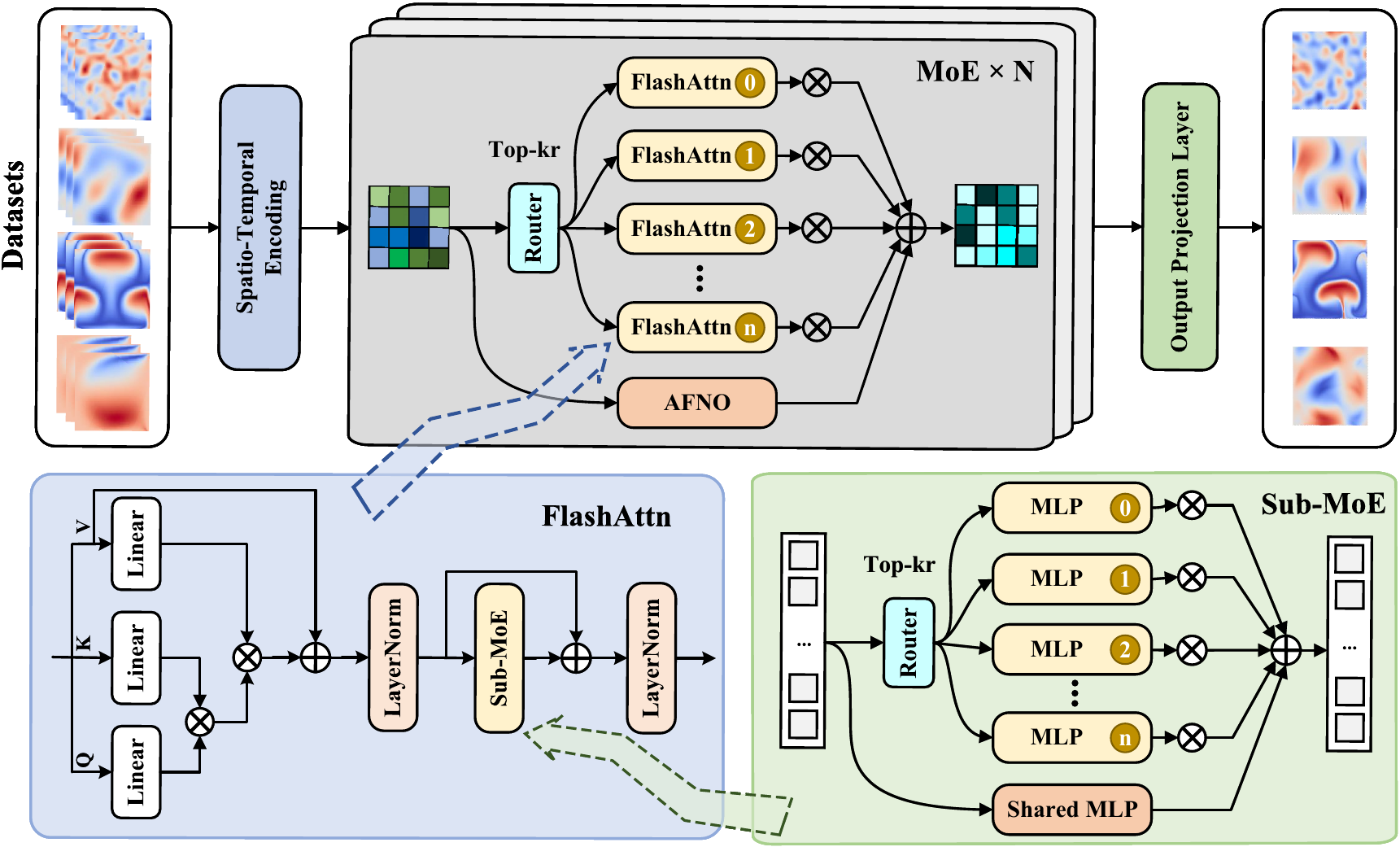}
    \caption{Overview architecture. We train on twelve mixed PDE datasets, predicting the next frame based on the preceding frames. We design a nested MoE architecture: (1) the top shows the overall model architecture; (2) the bottom right illustrates the nested Sub-MoE architecture; and (3) the bottom left depicts the improved Flash Attention architecture.}
    \label{fig:overview}
\end{figure*}

We propose a nested MoE framework (NESTOR), as shown in Fig.~\ref{fig:overview}. First, the PDE input is mapped to a latent representation space~\cite{hao2024dpot}. Then, these representations are input into nested MoE modules, and a gating mechanism assigns them to different experts, each capable of learning specific features of the input. The proposed network architecture integrates frequency domain features and spatiotemporal features, addressing the diversity and complexity of PDEs at both the image and token levels, thus demonstrating strong robustness and generalization ability.

\subsection{Spatio-Temporal Encoding}
First, the input $x \in \mathbb{R}^{B \times C \times H \times W}$ 
is divided into a set of non-overlapping patches 
$X_p \in \mathbb{R}^{B \times N \times C \times P_H \times P_W}$, 
where $B$ is the batch size, $N$ is the number of patches, and $(P_H \times P_W)$ is the patch size. 
Each patch is then projected into a $D$-dimensional space, followed by the addition of positional encoding $E_\text{pos}$~\cite{dosovitskiy2020image}:
\begin{equation}
X = \text{Embedding}(X_p) + E_\text{pos} \in \mathbb{R}^{B \times N \times D},
\end{equation}
Subsequently, the obtained representation is rearranged as 
$X \in \mathbb{R}^{B \times X \times Y \times T \times C}$, 
and mapping time series to a fixed dimension to compress information in the time dimension:  
\vspace{-6pt}
\begin{equation}
Y = \sum_{t=1}^{T} W_t X_t, \quad 
Y \in \mathbb{R}^{B \times X \times Y \times C_\text{out}},
\vspace{-6pt}
\end{equation}
where $W \in \mathbb{R}^{T \times C_\text{out} \times C_\text{out}}$ is a learnable weight matrix.

\subsection{Nested Mixture-of-Experts Architecture}
A single type of network architecture is insufficient to fully capture the diverse characteristics of data. To address this, we introduce a nested MoE architecture at the operator level to enable multi-scale interactions within the PDE system. This module dynamically allocates the most appropriate expert network through a routing mechanism, allowing it to simultaneously characterize both local and global dependencies and effectively capture features in both the time and frequency domains. Here, both the image-level MoE and the token-level Sub-MoE consist of 6 non-shared experts and 1 shared expert, with the gating network activating 2 of the non-shared experts.

\subsubsection{Image-level MoE}
\textbf{Routing Strategy.} We adopt an image-level gating mechanism and employ a top-k routing strategy~\cite{shazeer2017outrageously} for expert selection. First, given the input feature $x \in \mathbb{R}^{B \times C \times H \times W}$, we apply global average pooling to obtain the image-level representation $\bar{x}_b \in \mathbb{R}^C$, where $b=1,\dots,B$.
Next, the image-level representation is fed into a learnable linear layer to produce the raw expert scores:
\begin{equation}
s_b = \bar{x}_b W^\top + b \in \mathbb{R}^N,
\end{equation}
where $W \in \mathbb{R}^{N \times C}$ is the expert weight matrix, $b \in \mathbb{R}^N$ is the bias term, and $N$ denotes the number of experts.
The raw scores are then normalized using the softmax to obtain the routing probabilities
\vspace{-6pt}
\begin{equation}
p_b = \texttt{softmax}(s_b), \quad \sum_{i=1}^{N} p_{b,i} = 1.
\vspace{-6pt}
\end{equation}
Finally, according to the top-$k$ routing strategy, the $k$ experts with the highest probabilities are selected. Let $\mathcal{I}_b$ denote the index set of the selected experts. For each selected expert $i \in \mathcal{I}_b$, the final routing weight is defined as:
\begin{equation}
w_{b,i} = \frac{p_{b,i}}{\sum_{j \in \mathcal{I}b} p{b,j}}, \quad i \in \mathcal{I}_b.
\end{equation}
\vspace{-6pt}

\noindent\textbf{Expert Design.} We select AFNO~\cite{guibas2021adaptive}~\cite{hao2024dpot} as the shared expert, responsible for capturing global low-frequency spatial features. First, the input feature $x \in \mathbb{R}^{B \times C \times H \times W}$ is Fourier transformed:
$\hat{x} = \mathcal{F}(x), \quad \hat{x} \in \mathbb{C}^{B \times H \times W \times C}$, where $\mathcal{F}(\cdot)$ is the FFT operation. Next, a complex convolution operation is performed in the frequency domain:
\vspace{-1pt}
\begin{align} 
&\hat{y}_{\mathrm{real}} = \sigma \Big( \hat{x}_{\mathrm{real}} W_1^{(r)} - \hat{x}_{\mathrm{imag}} W_1^{(i)} + b_1^{(r)} \Big),\\
&\hat{y}_{\mathrm{imag}} = \sigma \Big( \hat{x}_{\mathrm{imag}} W_1^{(r)} + \hat{x}_{\mathrm{real}} W_1^{(i)} + b_1^{(i)} \Big),
\vspace{-5pt}
\end{align}
where $\sigma$ is the activation function, $W_1^{(r)}, W_1^{(i)}$ are learnable matrices for the real and imaginary parts, respectively, and $b_1^{(r)}, b_1^{(i)}$ are bias terms. Then, an inverse Fourier transform is performed to return to the spatiotemporal representation:
\begin{equation}
y = \mathcal{F}^{-1}(\hat{y}),
\end{equation}
where $\mathcal{F}^{-1}(\cdot)$ represents the IFFT operation. Finally, a normalization layer, MLP, and residual connections are combined to obtain the output of the shared expert.

In addition, we introduce Flash Attention~\cite{dao2022flashattention} as a non-shared expert, responsible for capturing the fine-grained spatiotemporal features of the physical field. First, the input feature \(x \in \mathbb{R}^{B \times C \times H \times W}\) is reshaped into a sequence form 
\(x' \in \mathbb{R}^{B \times C \times N}\) , where $\ N$ = $H \times W$. Next, \(x'\) is normalized and linearly transformed to obtain the query (\(Q\)), key (\(K\)), and value (\(V\)) representations. The attention-weighted result is then computed as
$Z = \text{softmax}\Big(\tfrac{Q K^\top}{\sqrt{d_k}}\Big) V$, which is added to the input residual and further normalized to obtain 
\(\tilde{Z}\). Subsequently, \(\tilde{Z}\) is passed through a Sub-MoE module for linear transformation:
\begin{equation}
Y = \texttt{Sub-MoE}(\tilde{Z}).
\end{equation}
Finally, by combining residual connections and normalization layers, we obtain the output of the non-shared expert.

\subsubsection{Token-level Sub-MoE}
\textbf{Routing Strategy.} We adopt a token-level gating mechanism and employ a top-k routing strategy for expert selection. Unlike the image-level gating mechanism, the token-level gates expert scores for each individual token vector, enabling finer-grained expert selection.

\noindent\textbf{Expert Design.} Our Sub-MoE implements the functionality of the FFN layer in Flash Attention. Both the shared and non-shared experts adopt the same network structure, which is an MLP. Normalized features are fed into the Sub-MoE, where token-level routing assigns them to the most appropriate expert for processing, extracting fine-grained feature representations. The computational process is as follows.
\begin{equation}
\texttt{ExpertMLP}(x) = W_2 \, \sigma(W_1 x + b_1) + b_2,
\end{equation}
where $W_1 \in \mathbb{R}^{C \times (rC)}$, $W_2 \in \mathbb{R}^{(rC) \times C}$, $r$ is $\text{mlp\_ratio}$, $\sigma(\cdot)$ denotes the activation function of GELU. Specifically, we first perform the first-layer linear transformation on the input feature $h = x W_1 + b_1$. Next, perform a nonlinear activation on $h$: $a = \texttt{GELU}(h)$. Finally, a second linear transformation is performed to obtain the final feature representation: $y = a W_2 + b_2$.

\subsection{Head and Loss Function}
\subsubsection{Load Balancing Loss}
In our nested MoE model, the routing mechanism assigns tokens to the most suitable experts. A balanced distribution of tokens among experts is crucial for MoE performance. When the allocation is imbalanced, some experts remain idle and fail to learn diverse features, while a few experts become overloaded, potentially causing memory bottlenecks. This can lead the model to degenerate to using only a subset of experts, failing to fully leverage the advantages of MoE~\cite{shazeer2017sparsely}. To address this issue, we introduce a load-balancing loss~\cite{shazeer2017outrageously} to encourage a more uniform distribution of tokens across experts. Here, the two load balancing losses are defined following the same pattern:
\vspace{-6pt}
\begin{equation}
\mathcal{L}_{\mathrm{aux}} = E \sum_{i=1}^E p_i \cdot f_i, 
\end{equation}
where $p_i = \frac{1}{N} \sum_{j=1}^N P_{ij}$ is the routing probability of expert $i$, $f_i = \frac{n_i}{\sum_{k=1}^E n_k}$ is the actual token assignment ratio of expert $i$, $E$ is the total number of experts, $N$ is the total number of tokens, $P_{ij}$ is the probability of token $j$ being assigned to expert $i$, and $n_i$ denotes the number of tokens assigned to expert $i$.

\subsubsection{Main Task Loss}
For our prediction task, we choose the relative error (L2RE)~\cite{li2020fourier} as the main task loss function:
\begin{equation}
\mathcal{L}_{2} = 
\tfrac{\left\| \hat{y}_i^{(c)} - y_i^{(c)} \right\|_{2}}
     {\left\| y_i^{(c)} \right\|_{2}},
\vspace{-6pt}
\end{equation}
where $y_i^{(c)}$ is the ground-truth of $i$-th sample at channel $c$, 
and $\hat{y}_i^{(c)}$ is the corresponding prediction.

\subsubsection{Total Loss}
Ultimately, our loss function consists of the main task loss and two load-balancing losses:
\begin{equation}
\mathcal{L} = \mathcal{L}_{2} + \alpha \mathcal{L}_{\mathrm{aux_1}} + \beta \mathcal{L}_{\mathrm{aux_2}},
\vspace{-3pt}
\end{equation}
where $\mathcal{L}_{2}$ denotes the main task's L2RE; $\mathcal{L}_{\mathrm{aux_1}}$ is the load balancing loss of Image-level MoE (image-level routing); $\mathcal{L}_{\mathrm{aux_2}}$ is the load balancing loss of the Image-level Sub-MoE (token-level routing); and $\alpha$ and $\beta$ are hyperparameters that control the contribution of the load balancing losses.

\section{Experiments} 

\newcolumntype{M}[1]{>{\centering\arraybackslash}m{#1}}

\begin{table*}[t]
\centering
\caption{The experiments are divided into two parts: one reports the pre-training performance of the model, and the other shows the fine-tuning results on each task. Here, “-200” denotes fine-tuning for 200 epochs, and “-500” for 500 epochs. The evaluation metric is L2RE. The best result within each part is highlighted in \textbf{bold}, while the overall best result is emphasized in \colorbox[RGB]{196,216,242}{blue bold.}}
\label{tab:main results}
\renewcommand{\arraystretch}{1.3}
\resizebox{\textwidth}{!}{%
\begin{tabular}{c|c|c|ccc|cccccccc|cc|c} 
\toprule
\toprule
 & L2RE & Activated & \multicolumn{3}{c|}{FNO-$\nu$} & \multicolumn{8}{c|}{PDEBench CNS-($\eta,\zeta$), DR, SWE} & \multicolumn{2}{c|}{PDEArena} & CFDBench \\
 & Model & Params & 1e-5 & 1e-4 & 1e-3 & 1,0.1 & 1,0.01 & Avg.(1) & 0.1,0.1 & 0.1,0.01 & Avg.(0.1) & DR & SWE & NS & NS-cond & - \\
\midrule
\multirow{9}{*}{\rotatebox[origin=c]{90}{\textbf{Pre-trained}}} &FNO        & 0.5M & 0.116 & 0.0922 & 0.0156 & 0.151 & 0.108 & 0.130 & 0.230 & 0.076 & 0.153 & 0.0321 & 0.0091 & 0.210 & 0.384 & 0.0274 \\
&UNet       & 25M & 0.198 & 0.119 & 0.0245 & 0.334 & 0.291 & 0.313 & 0.569 & 0.357 & 0.463 & 0.0971 & 0.0521 & 0.102 & 0.337 & 0.209 \\
&FFNO       & 1.3M & 0.121 & 0.0503 & 0.0099 & 0.0212 & 0.052 & 0.0366 & 0.162 & 0.0452 & 0.104 & 0.0571 & 0.0116 & 0.0839 & 0.602 & 0.0071 \\
&GK-T       & 1.6M & 0.134 & 0.0792 & 0.0098 & 0.0341 & 0.0377 & 0.0359 & 0.0274 & 0.0366 & 0.0320 & 0.0359 & 0.0069 & \textbf{0.0952} & 0.423 & 0.0105 \\
&GNOT       & 1.8M & 0.157 & \textbf{0.0443} & 0.0125 & 0.0325 & 0.0420 & 0.0373 & 0.0228 & 0.0341 & 0.0285 & 0.0311 & 0.0068 & 0.172 & \textbf{0.325} & \textbf{0.0088} \\
&Oformer    & 1.9M & 0.1705 & 0.0645 & 0.0104 & 0.0417 & 0.0625 & 0.0521 & 0.0254 & 0.0205 & 0.0229 & 0.0192 & 0.0072 & 0.135 & 0.332 & 0.0102 \\
&MPP-T      & 7M & - & - & - & - & - & 0.0442 & - & - & 0.0312 & \textbf{0.0168} & 0.0066 & - & - & - \\
&DPOT-T     & 7M & \textbf{0.0976} & 0.0606 & 0.00954 & 0.0173 & 0.0397 & 0.0285 & 0.0132 & 0.0220 & 0.0176 & 0.0321 & 0.0056 & 0.125 & 0.384 & 0.0095 \\
&\textbf{Ours}       & 13M & 0.1195 & 0.0951 & \textbf{0.0093} & \textbf{0.0167} & \textbf{0.0373} & \textbf{0.0270} & \colorbox[RGB]{196,216,242}{\textbf{0.0120}} & \textbf{0.0202} & \textbf{0.0161} & 0.0308 & \textbf{0.0052} & 0.132 & 0.409 & 0.0112 \\
\midrule
\multirow{4}{*}{\rotatebox[origin=c]{90}{\textbf{Fine-tune}}} &DPOT-FT200 & 7M & 0.0511 & 0.0431 & 0.0073 & 0.0136 & 0.0238 & 0.0187 & 0.0168 & 0.0145 & 0.0157 & 0.0194 & 0.0028 & 0.103 & 0.313 & 0.0054 \\
&\textbf{Ours}-FT200 & 13M & 0.0581 & 0.0313 & 0.0056 & 0.0139 & 0.0182 & 0.0161 & 0.0155 & 0.0112 & 0.0134 & 0.0198 & 0.0032 & 0.0793 & 0.321 & 0.0045 \\
&DPOT-FT500 & 7M & 0.0520 & 0.0367 & 0.0058 & 0.0112 & 0.0195 & 0.0153 & 0.0174 & 0.0138 & 0.0156 & 0.0148 & \colorbox[RGB]{196,216,242}{\textbf{0.0024}} & 0.0910 & \colorbox[RGB]{196,216,242}{\textbf{0.280}} & 0.0039 \\
&\textbf{Ours}-FT500 & 13M & \colorbox[RGB]{196,216,242}{\textbf{0.0505}} & \colorbox[RGB]{196,216,242}{\textbf{0.0217}} & \colorbox[RGB]{196,216,242}{\textbf{0.0043}} & \colorbox[RGB]{196,216,242}{\textbf{0.0094}} & \colorbox[RGB]{196,216,242}{\textbf{0.0134}} & \colorbox[RGB]{196,216,242}{\textbf{0.0114}} & \textbf{0.0123} & \colorbox[RGB]{196,216,242}{\textbf{0.0083}} & \colorbox[RGB]{196,216,242}{\textbf{0.0103}} & \colorbox[RGB]{196,216,242}{\textbf{0.0117}} & 0.0026 & \colorbox[RGB]{196,216,242}{\textbf{0.0683}} & 0.285 & \colorbox[RGB]{196,216,242}{\textbf{0.0038}} \\
\bottomrule
\end{tabular}%
}
\vspace{-10pt}
\end{table*}

\subsection{Datasets and Evaluation Metric} 
\textbf{Datasets.} We conduct experiments on a mixed dataset consisting of twelve different data sources and different parameters from FNO~\cite{li2020fourier}, PDEBench~\cite{takamoto2022pdebench}, PDEArena~\cite{gupta2022towards}, and CFDBench~\cite{luo2023cfdbench}. (1) FNO: A dataset containing three different parameters for the same type of equation. (2) PDEBench: A dataset containing four different parameters for the same type of equation. (3) PDEArena: A dataset containing the same equation with and without initial conditions. (4) CFDBench: A multi-task PDE dataset obtained by processing the four subtasks uniformly.

\noindent\textbf{Evaluation Metrics.} We use L2RE~\cite{li2020fourier} as the evaluation metric, where lower L2RE indicates better performance.

\subsection{Main Results} 
Table~\ref{tab:main results} presents the experimental results of our method compared with other models in the pre-training datasets. The first row of the table specifies the types of PDE datasets and parameter settings, while the first column lists the baseline models for comparison. The experiments are divided into two parts: the first is pre-training, where all models are trained from scratch on the datasets; the second is fine-tuning, where models are further trained based on the pre-trained weights.

In the pre-training stage, our method demonstrates strong performance across 12 PDE datasets, achieving state-of-the-art results on 6 of them. Notably, our model ranks first on 5 out of 6 PDEBench datasets, and achieves significantly lower errors than mainstream models on multiple benchmarks. These results clearly validate the effectiveness of our proposed architecture for handling complex PDE systems, highlighting its superior performance and generalizability in PDE modeling.

In the fine-tuning stage, we conduct 200 and 500 epochs of fine-tuning on each dataset. The results show that after 500 epochs, our model achieves state-of-the-art performance on 9 out of 12 tasks, surpassing advanced pre-trained models on the majority of tasks. Compared with training from scratch, fine-tuning on pretrained weights generally leads to better performance; moreover, increasing the number of fine-tuning steps typically yields higher prediction accuracy. These results demonstrate the superior performance of our proposed model on sparse datasets with stronger generalization and adaptability.

In summary, our model demonstrates significant advantages in operator learning for PDE tasks. With the aid of fine-tuning strategies, it can rapidly adapt to specific tasks and achieve a total of 10 global best performances across 12 benchmark datasets, highlighting its strong modeling capability in capturing complex dynamics and multi-scale features, as well as its excellent generalization ability.

\begin{figure}[h]
\includegraphics[width=01\linewidth]{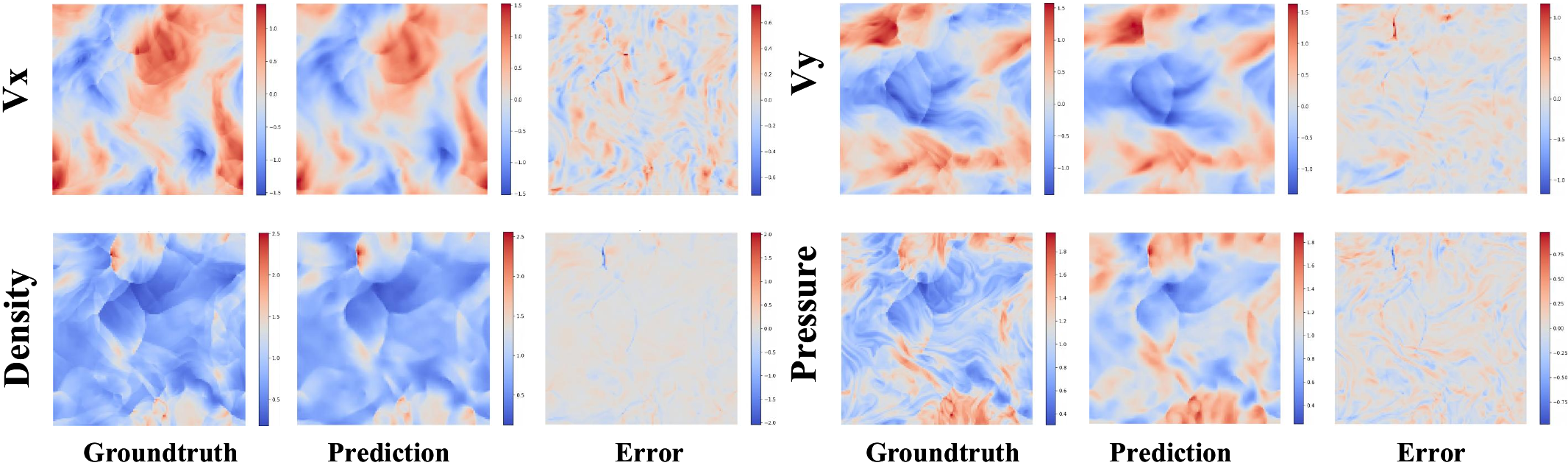}
\caption{Visualization of 2D high-resolution turbulence prediction results. (1) The first column shows the true values, the second shows the model predictions, and the third shows the corresponding errors. (2) The predicted physical quantities are horizontal velocity, vertical velocity, density field, and pressure field.}
\vspace{-6pt}
\label{fig:turb_vis}
\end{figure}

\begin{figure}[h]
\includegraphics[width=1\linewidth]{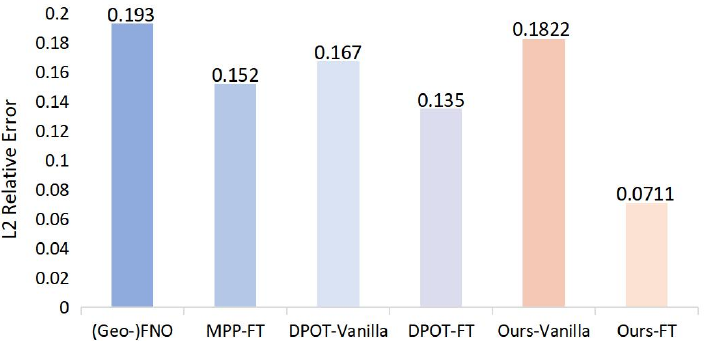}
\caption{Performance comparison of different models on the 2D high-resolution turbulence task. We use L2RE as the evaluation metric, where Vanilla denotes training from scratch, and -FT indicates results after 500 fine-tuning epochs on the downstream task.}
\vspace{-12pt}
\label{fig:turb}
\end{figure}

\subsection{Downstream Tasks Experiments} 
To evaluate the generalization and transferability of our model, we conduct downstream experiments on a two-dimensional high-resolution turbulence task ($ 512 \times 512$). In these experiments, we reuse most of the parameters from the pre-trained model, including the weights of the MoE modules and the spatio-temporal encoding. The visualization of the model predictions is shown in Fig.~\ref{fig:turb_vis}.

As illustrated in Fig.~\ref{fig:turb}, most models fine-tuned from pre-trained weights outperform those trained from scratch, which demonstrates the effectiveness of large-scale pre-training. This indicates that the model can acquire generalizable PDE knowledge and successfully transfer it to specific downstream tasks. On the two-dimensional high-resolution turbulence task, our model achieves a 47.3$\%$ improvement in prediction accuracy, reaching the best performance. The experimental results demonstrate that our pre-trained model learns more effective representations, achieving strong transfer performance on downstream tasks with only minimal fine-tuning. Moreover, it maintains precise prediction capability even on high-resolution tasks, fully showcasing its advantage in capturing PDE characteristics.

\begin{table}[t]
\centering
\caption{The impact of the number of non-shared experts. We use L2RE as the evaluation metric.}
\label{tab:scaling_1}
\resizebox{1\linewidth}{!}{
\begin{tabular}{ccccc@{\hspace{10pt}}r} 
\toprule
Setting & Num. of experts & FNO & PDEBench & SWE & Avg. $\mathcal{L}_{2}$ \\
\midrule
\multirow{4}{*}{FT-200}
 & 2  & 0.0575 & 0.0182 & 0.0024 & 0.0262 \\
 & 4  & 0.0577 & 0.0240 & 0.0579 & 0.0466 \\
 & 6  & 0.0563 & 0.0150 & 0.0025 & 0.0246 \\
 & 12 & 0.0575 & 0.1896 & 0.0025 & 0.0832 \\
\midrule
\multirow{4}{*}{FT-500}
 & 2  & 0.0519 & 0.0126 & 0.0022 & 0.0222 \\
 & 4  & 0.0504 & 0.0114 & 0.0025 & 0.0214 \\
 & 6  & 0.0502 & 0.0115 & 0.0021 & 0.0213 \\
 & 12 & 0.0520 & 0.0144 & 0.0025 & 0.0230 \\
\bottomrule
\end{tabular}
}
\end{table}

\begin{table}[t]
\centering
\caption{The impact of the size of the pre-training data on performance. We use L2RE as the evaluation metric.}
\label{tab:scaling_2}
\resizebox{1\linewidth}{!}{
\begin{tabular}{ccccc}
\toprule
Num. of datasets & FNO & PDEBench & SWE & Avg. $\mathcal{L}_{2}$ \\
\midrule
3  & 0.0512 & 0.0165 & 0.0026 & 0.0234 \\
12 & 0.0505 & 0.0094 & 0.0026 & 0.0208 \\
\bottomrule
\end{tabular}
}
\vspace{-12pt}
\end{table}

\subsection{Scaling Experiments} 
The number of experts in the MoE architecture is a key factor influencing the performance of pre-trained models. Under the setting where the number of activated experts per forward pass is fixed, we vary the number of unshared experts and use the average L2RE across datasets as the evaluation metric to study how the number of non-shared experts affects pre-training performance. On the selected datasets, we adopt two fine-tuning strategies: FT-200 (200 steps of fine-tuning) and FT-500 (500 steps of fine-tuning). As shown in Table~\ref{tab:scaling_1}, the results indicate that fine-tuning the pre-trained model significantly improves task performance, and additional fine-tuning steps lead to further gains. For complex MoE architectures, however, having more experts is not always better; increasing the number of experts makes optimization more challenging and complicates resource allocation. For different tasks, there typically exists an optimal range for the number of experts, and selecting an appropriate expert size is essential for fully realizing the performance potential of MoE models.

We investigate the impact of pre-training data scale on model performance, as shown in Table~\ref{tab:scaling_2}. Specifically, we conduct pre-training on 3 and 12 different PDE datasets, followed by 500 epochs of fine-tuning on each downstream task. The results demonstrate that increasing the amount of pre-training data improves fine-tuning performance, indicating that large-scale cross-equation pre-training effectively enhances the model’s generalization capability.


\begin{table}[t]
\centering
\small 
\caption{Ablation experiments of our proposed model on the PDEBench datasets. “w/o” denotes the removal of the corresponding component. We use L2RE as the evaluation metric.}
\label{tab:3}
\resizebox{\columnwidth}{!}{
\begin{tabular}{lcccccccc}
\toprule
Method & 1,0.1 & 1,0.01 & 0.1,0.1 & 0.1,0.01 & DR & SWE & Avg. $\mathcal{L}_{2}$ & Promotion \\
\midrule
Ours & 0.0144 & 0.0355 & 0.0135 & 0.0178 & 0.0282 & 0.0045 & 0.0173 & \textbf{-} \\
w/o Sub-MoE & 0.0157 & 0.0393 & 0.0130 & 0.0209 & 0.0245 & 0.0049 & 0.0197 & \textbf{0.0024} \\
w/o Load Balance Loss & 0.0135 & 0.0335 & 0.0109 & 0.0159 & 0.0265 & 0.0062 & 0.0178 & \textbf{0.0005} \\
FlashAttn + AFNO Sum & 0.0149 & 0.0363 & 0.0136 & 0.0178 & 0.0304 & 0.0046 & 0.0196 & \textbf{0.0023} \\
\bottomrule
\end{tabular}}
\vspace{-6pt} 
\end{table}

\subsection{Ablation Studies} 
To validate the effectiveness of our model, we conduct experiments on six sub-tasks of the PDEBench dataset to assess the impact of different modules on model performance. Using the complete model as the baseline, we systematically perform ablation studies by progressively removing or replacing key modules, with the average L2RE (Avg. $\mathcal{L}_{2}$) serving as the primary comprehensive evaluation metric. The results are shown in Table~\ref {tab:3}.

\noindent\textbf{Impact of Sub-MoE.} Removing the Sub-MoE module leads to an increase of 0.0024 in the average L2RE. Among all modules, Sub-MoE contributed most significantly to performance improvement, indicating that it plays an important role in effectively capturing multi-scale and diverse features, thereby fully validating its importance.

\noindent\textbf{Impact of the Load Balancing Loss.} Removing the load balancing loss results in an increase of 0.0005 in the average L2RE. Although its contribution is smaller compared to other modules, it still provides a certain improvement to model performance.

\noindent\textbf{Impact of the Fusion Strategy.} Changing the fusion of AFNO and FlashAttention from MoE to simple addition increases the Avg. L2RE by 0.0023. This demonstrates that our model can select the most suitable experts for different inputs, thereby enhancing model performance and generalization ability, and validates the rationality of the design.

\subsection{Interpretable Analysis} 
To verify the effectiveness of the proposed nested MoE (Mixture of Experts) architecture, we conduct experiments at two levels: the global feature modeling capability of image-level experts and the local region modeling capability of token-level experts.

\noindent\textbf{Effectiveness of Image-Level MoE.} The image-level gating network generates expert scores based on the global features of the input samples and activates the two experts with the highest scores through a Top-2 selection mechanism. We statistically analyze the activation frequency of each expert on different PDE-type datasets to examine the correlation between expert selection and equation type. The results are shown in Table~\ref{tab:expert_dist}. It can be seen that Expert 0 and Expert 1 show a significant preference in the NS2D (Navier–Stokes) dataset, with a combined activation rate exceeding 70\%, indicating that these two experts are adept at handling complex flow characteristics dominated by convection. Expert 2 and Expert 3 dominate activation on the SWE (Shallow Water Wave Equation) dataset, with a combined activation frequency of 99.74\%, demonstrating their ability to model the characteristics of wave propagation processes. Expert 0 and Expert 5 perform outstandingly in the DR (Diffusion–Reaction) dataset, with a combined activation rate of 78.77\%, indicating their ability to capture the chemical reaction source term and diffusion-flow coupling effect in the diffusion process. In the two similar equation datasets with different parameters, M1(-1,-1) and M-1(-1,-1), Expert 0 and Expert 1 are also frequently selected, indicating that image-level MoE can effectively distinguish PDE types and select the optimal expert combination. Experimental results show that image-level experts can adaptively identify global features of different PDE types and automatically select the most suitable expert combination for modeling through a gating mechanism. 

\begin{figure}
    \centering
    \includegraphics[width=1\linewidth]{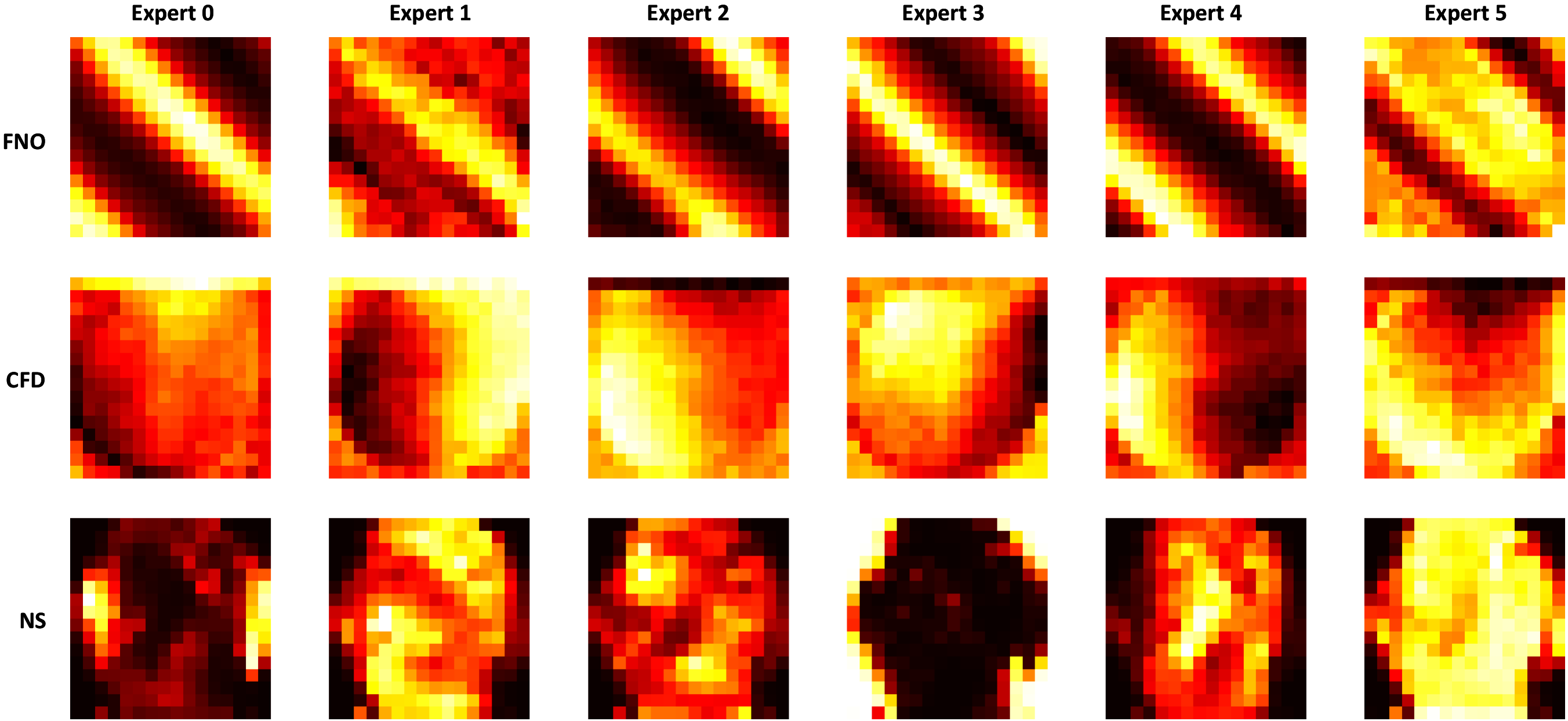}
    \caption{Visualization of spatial activation patterns produced by token-level experts in the Sub-MoE layer. For each input sample, the expert activation probabilities are projected onto heatmaps.}
    \label{fig:vis_moe2}
    \vspace{-12pt}
\end{figure}

\noindent\textbf{Effectiveness of Token-Level MoE.} To verify the spatial region modeling capability of token-level experts, we conduct a visualization experiment based on spatial heatmaps. For each input sample, the activation probabilities of token-level experts are extracted from the Sub-MoE layer, generating a heatmap, as shown in Fig.~\ref{fig:vis_moe2}. The visualization results show that different token-level experts exhibit distinct region-specific activation patterns in space. This pattern indicates that token-level MoE can effectively capture local region correlations within the physical field, providing a more refined expressive capability for modeling complex multi-scale physical systems.

In summary, our nested MoE architecture is effective. At the macroscopic level, image-level experts achieve adaptive functional division based on PDE types; at the microscopic level, token-level experts effectively capture regional correlations within the physical field. This dual specialization mechanism of "macroscopic classification – microscopic partitioning" significantly improves the model's modeling and generalization capabilities for complex multiphysics problems.

\begin{table}[t]
\centering
\caption{Expert selection distribution of the MoE router across different datasets. The top two experts for each dataset are highlighted in \textbf{bold}(\%).}
\label{tab:expert_dist}
\resizebox{\linewidth}{!}{
\begin{tabular}{lcccccc}
\toprule
Dataset & Expert 0 & Expert 1 & Expert 2 & Expert 3 & Expert 4 & Expert 5 \\
\midrule
M1(-1,-1)   & \textbf{22.39} & \textbf{50.00} & 6.98 & 5.37 & 10.69 & 4.58 \\
M-1(-1,-1)  & \textbf{22.88} & \textbf{50.00} & 3.51 & 11.58 & 10.22 & 1.81 \\
SWE         & 0.00  & 0.00  & \textbf{50.00} & \textbf{49.74} & 0.00 & 0.25 \\
DR          & \textbf{28.77} & 0.00  & 2.31 & 12.64 & 6.28 & \textbf{50.00} \\
\bottomrule
\end{tabular}}
\vspace{-6pt} 
\end{table}

\begin{table}[t]
\centering
\caption{Comparison of activation parameters, total parameters, and activation rates of different models.}
\label{tab:params}
\resizebox{1\linewidth}{!}{
\begin{tabular}{lccc}
\toprule
 & ~~~DPOT-T~~~ & ~~~MoE-POT-T~~~ & ~~~Ours~~~ \\
\midrule
Activated Parameters & 7.5M  & 17M  & 13M \\
Total Parameters     & 7.5M  & 30M  & 83M \\
Activation Ratio     & 100\% & 56.67\% & 16.67\% \\
\bottomrule
\end{tabular}}
\vspace{-12pt}
\end{table}

\subsection{Efficiency Analysis} 
We analyze the efficiency of our models, as shown in Table~\ref{tab:params}. Traditional single-network architectures, such as DPOT-T, can only increase model capacity by adding more parameters, which typically leads to a linear growth in computational cost. In contrast, models incorporating the Mixture-of-Experts (MoE) mechanism, such as MoE-POT-T~\cite{wang2025mixture} and Ours, can expand model capacity through selective activation of expert sub-networks, thereby improving performance while keeping computational costs low.

Specifically, although ours has a much larger total number of parameters compared to DPOT-T and MoE-POT-T, its activated parameter ratio is only 16.67\%, significantly lower than MoE-POT-T’s 56.67\% and DPOT-T’s 100\%. This demonstrates that the selective activation of MoE not only allows the model to achieve higher capacity without increasing the actual computational burden but also provides a practical solution for efficient scaling.

\section{Conclusion} 
This paper proposes a large-scale PDE pre-trained neural operator based on a nested Mixture-of-Experts (MoE) architecture. We design the nested MoE framework, which consists of image-level MoE and token-level MoE, and conduct extensive training on twelve PDE datasets to obtain a universal pre-trained model. Our model successfully transfers to specific tasks and new downstream tasks, achieving state-of-the-art performance on most datasets. Furthermore, this paper explores the suitability and advantages of MoE architectures for large-scale PDE pre-trained neural operators, pioneers the design of a hierarchical MoE architecture in this field, and reveals new potential for solving PDEs.

{
    \small
    \bibliographystyle{ieeenat_fullname}
    \bibliography{main}
}

\clearpage

\appendix

\section{appendix}
\subsection{LLM USAGE}
During the manuscript writing and revision process, we used a Large Language Model (LLM) to assist. Specifically, LLM was used to improve the accuracy and readability of the language, and to help ensure the overall structure and clarity of the paper. This tool primarily assisted with tasks such as sentence reconstruction, grammatical proofreading, and improving text coherence.

\subsection{Experimental Details}
\textbf{Pre-training.} We pre-trained the model on 8 NVIDIA RTX 4090 GPUs using the Adam optimizer with an initial learning rate of $1.0 \times 10^{-3}$ and a cyclic learning rate schedule (cycle), including 200 warm-up epochs. The total training lasted 1000 epochs with a batch size of 32. To mitigate the effects of varying dataset sizes, training weights were assigned to each dataset. During training, we used $T=10$ time steps to predict the next frame, maintaining consistency with the original settings of most datasets. The details are shown in Table~\ref{tab:setting}.

\noindent\textbf{Fine-tuning.} In the fine-tuning stage, we loaded the pre-trained weights and performed 200-epoch and 500-epoch fine-tuning on each subset. The key module of the model is the nested MoE layer, whose parameters are shared across different frequency components along the channel dimension, enabling cross-level expert collaboration.

\begin{table*}[t]
\centering
\caption{Setting of the Attention Module.}
\resizebox{1\textwidth}{!}{%
\begin{tabular}{cccccccccccc}
\toprule
\textbf{Dim} & \textbf{Ratio} & \textbf{Layers} & \textbf{Heads} & \textbf{Routed$_1$} & \textbf{Shared$_1$} & \textbf{Top-$k_1$} & \textbf{Routed$_2$} & \textbf{Shared$_2$} & \textbf{Top-$k_2$} & \textbf{Model Size} & \textbf{Activated Size} \\
\midrule
512 & 1 & 2 & 4 & 1 & 6 & 2 & 1 & 6 & 2 & 83M & 13M \\
\bottomrule
\end{tabular}%
}
\label{tab:setting}
\end{table*}

\subsection{Data Preprocessing and Sampling}
\textbf{Data Padding and Masking.} Different PDE datasets vary in resolution, number of variables, and geometric configurations. If we directly sample from the raw data, the resulting batch will have large variations in size, leading to unbalanced training loads and reduced efficiency in modern multi-GPU training. Here, we adopt the padding and masking strategy from DPOT. First, we select a fixed resolution $H = 128$, which matches a considerable portion of the datasets. Datasets with lower resolutions are upsampled to $H$ via interpolation, while those with higher resolutions are randomly downsampled or interpolated to $H$. Second, to unify the number of variables across different PDEs, we pad all datasets along the channel dimension (e.g., filling with ones) to match the maximum number of channels. For datasets with irregular geometries, an additional mask channel is introduced to encode the specific geometric configuration of each PDE instance.

\noindent\textbf{Balanced Data Sampling.} When training with multiple PDE datasets, differences among datasets can lead to unbalanced training progress and inefficiency. To address this issue, we adopt the sampling strategy from DPOT, which balances the sampling probabilities across datasets during training. Our goal is to ensure that each dataset is represented equally throughout the training process. Let $|D_k|$ denote the number of samples in the $k$-th dataset ($1 \le k \le K$), and assign a weight $w_k$ to each dataset to indicate its relative importance. Then, the sampling probability from dataset $D_k$ is defined as:
\[
p_k = \frac{w_k}{K\,|D_k|\,\sum_{k} w_k}
\]
We can observe that the sampling probability depends on the weight $w_k$ rather than the dataset size $|D_k|$, which helps mitigate gradient imbalance caused by dataset size disparities.

\subsection{Limitations and Conclusions}
We use DPOT as our primary baseline and adopt its data processing strategies, including adding noise, data padding, and balanced data sampling. However, our core model differs from DPOT, which is based on AFNO, while our network architecture employs a nested MoE. MoE-POT, our work, also incorporates a MoE architecture, but uses only a single-layer MoE and primarily improves upon the frequency convolution in AFNO. In contrast, our proposed nested MoE architecture processes PDE features at both macroscopic and microscopic levels, achieving an effective fusion of frequency domain and spatiotemporal domain features.

Due to resource constraints, our model is currently only implemented with one parameter size, but this version has already demonstrated good accuracy and generalization ability. Combining the results of scaling experiments and interpretability analysis, we validate the model's effectiveness and show that it can be scaled to versions with different parameter sizes. Considering the diversity of expert and activation numbers, future work can explore optimal parameter configurations to further improve model performance.

\begin{table*}[t]
\caption{Train and test set sizes of the PDE datasets used for pre-training.}
\label{tab:5}
\centering
\renewcommand{\arraystretch}{1.3}
\resizebox{\textwidth}{!}{%
\begin{tabular}{c|ccc|cccccc|cc|c}
\toprule
 & \multicolumn{3}{c|}{FNO-$\nu$} & \multicolumn{6}{c|}{PDEBench CNS-($\eta,\zeta$), DR, SWE} & \multicolumn{2}{c|}{PDEArena} & CFDBench \\
 & 1e-5 & 1e-4 & 1e-3 & 1,0.1 & 1,0.01 & 0.1,0.1 & 0.1,0.01 & DR & SWE & NS & NS-cond & - \\
\midrule
Train set size & 100 & 9800 & 1000 & 9000 & 9000 & 9000 & 9000 & 900 & 900 & 6500 & 3100 & 9000 \\
\hline
Test set size & 200 & 200 & 200 & 1000 & 1000 & 1000 & 1000 & 100 & 100 & 1300 & 600 & 1000 \\
\bottomrule
\end{tabular}%
}
\end{table*}

\subsection{Detailed Information of Datasets}
We list the configurations of the PDE datasets used for pre-training along with detailed descriptions of the governing partial differential equations:

\noindent\textbf{FNO-$v$}: This dataset focuses on the temporal evolution of the two-dimensional incompressible fluid vorticity field $w(x,t)$, where $(x,t) \in [0,1]^2 \times [0,T]$. The dynamics are governed by the two-dimensional Navier--Stokes equations in the vorticity--streamfunction formulation:
\begin{equation}
    \partial_t w + u \cdot \nabla w = \nu \Delta w + f(x), 
    \qquad \nabla \cdot u = 0,
\end{equation}
where $u$ denotes the velocity field, $\nu$ is the viscosity coefficient, $\Delta$ represents the Laplace operator, and $f(x)$ denotes the external forcing term. By varying the viscosity $\nu$, the dataset provides fluid dynamics simulations under different flow regimes, enabling the study of how viscosity influences the evolution of vortex structures.

\noindent\textbf{PDEBench-CMS}: This dataset focuses on the numerical simulation of compressible fluid mechanics (CMS). The goal is to predict the temporal evolution of the velocity field $u(x,t)$, the pressure field $p(x,t)$, and the density field $\rho(x,t)$ over the spatio-temporal domain $(x,t) \in [0,1]^2 \times [0,1]$. The data are generated based on the governing equations of compressible fluid dynamics, which consist of the conservation of mass, momentum, and energy:
\begin{equation}
\partial_t \rho + \nabla \cdot (\rho u) = 0,
\end{equation}
\begin{equation}
\rho \left( \partial_t u + u \cdot \nabla u \right) = - \nabla p + \eta \Delta u + \left( \zeta + \tfrac{\eta}{3} \right) \nabla (\nabla \cdot u),
\end{equation}
\begin{equation}
\partial_t \left( \tfrac{3}{2}p + \tfrac{\rho u^2}{2} \right)
= - \nabla \cdot \Big[ \Big(\varepsilon + p + \tfrac{\rho u^2}{2}\Big)u - u \cdot \sigma' \Big],
\end{equation}
where $\eta$ denotes the shear viscosity coefficient and $\zeta$ the bulk viscosity coefficient.$\varepsilon$ is the energy density and $\sigma'$ is the stress tensor.

\noindent\textbf{PDEBench-SWE}: The dataset is derived from PDEBench and focuses on the numerical simulation of the Shallow Water Equations (SWE). The objective is to predict the water depth field $h(x,t)$ over the spatiotemporal domain $(x,t) \in [-1,1]^2 \times [0,5]$. The SWE is a set of approximate governing equations widely used in ocean dynamics, flood modeling, and geomorphological evolution studies. The governing equations are given as follows:
\begin{equation}
\partial_t h + \nabla \cdot (h u) = 0,
\end{equation}
\begin{equation}
\partial_t (h u) + \nabla \cdot \left( \tfrac{1}{2} h u^2 + \tfrac{1}{2} g r h^2 \right) = - g r h \nabla b,
\end{equation}

\noindent\textbf{PDEBench-DR}: The dataset is derived from PDEBench and focuses on the numerical simulation of diffusion–reaction (DR) systems. The objective is to predict the density field $u(x,t)$ over the spatiotemporal domain $(x,t) \in [-2.5, 2.5]^2 \times [0,1]$. The governing equation is given by:
\begin{equation}
\partial_t u = D \nabla^2 u + R(u),
\end{equation}
where $D$ is the diffusion coefficient and $R(u)$ denotes the nonlinear reaction term.

\noindent\textbf{PDEArena}: The dataset is derived from PDEArena and focuses on the numerical simulation of incompressible Navier–Stokes (NS) flows. The objective is to predict the velocity field $u(x,t)$, pressure field $p(x,t)$, and density field $\rho(x,t)$ over the spatiotemporal domain $(x,t) \in [0,32]^2 \times [0,24]$. 

The 2D incompressible Navier--Stokes equations are given by:
\begin{equation}
\frac{\partial \mathbf{u}}{\partial t} + (\mathbf{u} \cdot \nabla) \mathbf{u} = - \nabla p + \nu \Delta \mathbf{u},
\end{equation}
\begin{equation}
\nabla \cdot \mathbf{u} = 0,
\end{equation}
where $\mathbf{u} = (u,v)^\top$ is the velocity field, $p$ is the pressure, and $\nu$ is the kinematic viscosity. 

NS-cond introduces additional physical conditions such as forcing fields $\mathbf{f}(\mathbf{x},t)$ or spatially varying viscosity $\nu(\mathbf{x})$:
\begin{equation}
\frac{\partial \mathbf{u}}{\partial t} + (\mathbf{u} \cdot \nabla) \mathbf{u} = - \nabla p + \nu(\mathbf{x}) \Delta \mathbf{u} + \mathbf{f}(\mathbf{x},t),
\end{equation}
\begin{equation}
\nabla \cdot \mathbf{u} = 0.
\end{equation}
Here, $\mathbf{f}(\mathbf{x},t)$ denotes external forcing and $\nu(\mathbf{x})$ can vary spatially.

\noindent\textbf{CFDBench}: The dataset is derived from CFDBench and focuses on the numerical simulation of incompressible or weakly compressible flows in irregular geometries. The objective is to predict the velocity field $u(x,t)$ and the pressure field $p(x,t)$ over domains with complex boundaries. The governing equations are given as follows:
\begin{equation}
\partial_t (\rho u) + \nabla \cdot (\rho u^2) = - \nabla p + \nabla \cdot \mu (\nabla u + \nabla u^\top),
\end{equation}
\begin{equation}
\nabla \cdot (\rho u) = 0,
\end{equation}
where $\rho$ is the fluid density, $u$ is the velocity field, $p$ is the pressure, and $\mu$ denotes the viscosity coefficient.

\subsection{Open access to data and code}
To ensure reproducibility, our code will be released upon acceptance of the paper. The experiments are conducted on publicly available datasets.

\begin{table}[t]
\centering
\caption{Comparison with MoE-POT in pre-training across six datasets. The evaluation metric is L2RE. The best result within each part is highlighted in \textbf{bold}.}
\label{tab:supplementary experiments}
\renewcommand{\arraystretch}{1.3}
\resizebox{1\columnwidth}{!}{%
\begin{tabular}{c|c|cc|ccc|c} 
\toprule
 L2RE & Activated & \multicolumn{2}{c|}{FNO-$\nu$} & \multicolumn{3}{c|}{PDEBench} & CFDBench \\
 Model & Params & 1e-5 & 1e-3 & 0.1,0.01 & SWE & DR & - \\
\midrule
MoE-POT  & 17M & 0.0682 & 0.00768 & \textbf{0.0105} & 0.00640 & 0.0411 & \textbf{0.00529} \\
Ours     & 13M & \textbf{0.0674} & \textbf{0.00763} & 0.0159 & \textbf{0.00449} & \textbf{0.0184} & 0.00911 \\
\bottomrule
\end{tabular}%
}
\vspace{-5pt}
\end{table}

\subsection{Supplementary Experiments} We compare our proposed model with the recently released MoE-based architecture MoE-POT in a mixed pre-training setting comprising six datasets, as shown in Table ~\ref{tab:supplementary experiments}. As can be seen, our model achieves new state-of-the-art results on four out of the six datasets, demonstrating its strong cross-equation generalization ability and unified modeling capability.

\subsection{Visualization} For each specific subtask, we first load the model weights pretrained on large-scale PDE datasets, and then fine-tune the model for the subtask. During fine-tuning, the model can adapt to the data distribution and equation characteristics of each subtask. The visualization of the prediction results is shown in the figure. For each data series, we select a representative equation to illustrate the model’s performance across different tasks. These visualizations allow us to observe the model’s ability to capture spatiotemporal trends, local details, and global patterns, thereby demonstrating the effectiveness and advantages of the pretrained weights in downstream tasks.

\begin{figure}[t]
    \centering
    \includegraphics[width=0.75\linewidth]{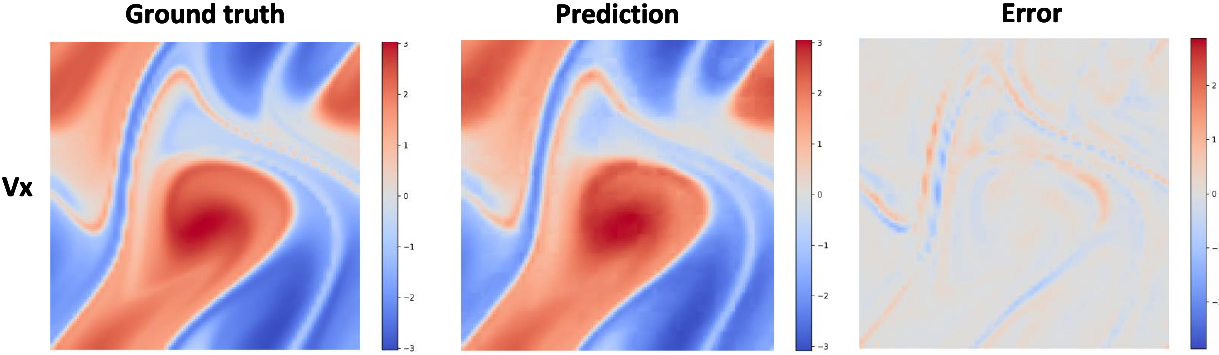}
    \caption{FNO series of result visualizations. (1) The first column shows the true value, the second column shows the model prediction value, and the third column shows the corresponding error. (2) Each row is the predicted physical quantity.}
    \label{fig:fno}
    \vspace{-12pt}
\end{figure}

\begin{figure}[t]
    \centering
    \includegraphics[width=0.75\linewidth]{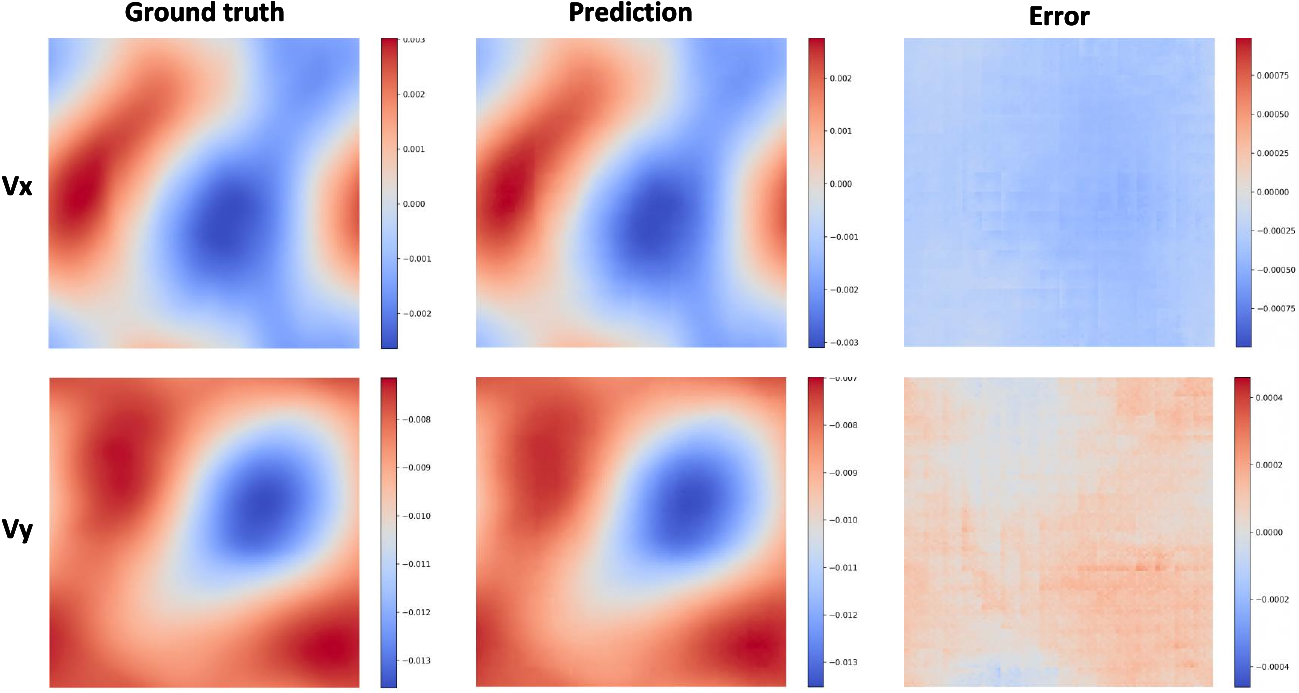}
    \caption{PDEBench series of result visualizations. (1) The first column shows the true value, the second column shows the model prediction value, and the third column shows the corresponding error. (2) Each row is the predicted physical quantity.}
    \label{fig:pdebench}
    \vspace{-12pt}
\end{figure}

\begin{figure}[t]
    \centering
    \includegraphics[width=0.75\linewidth]{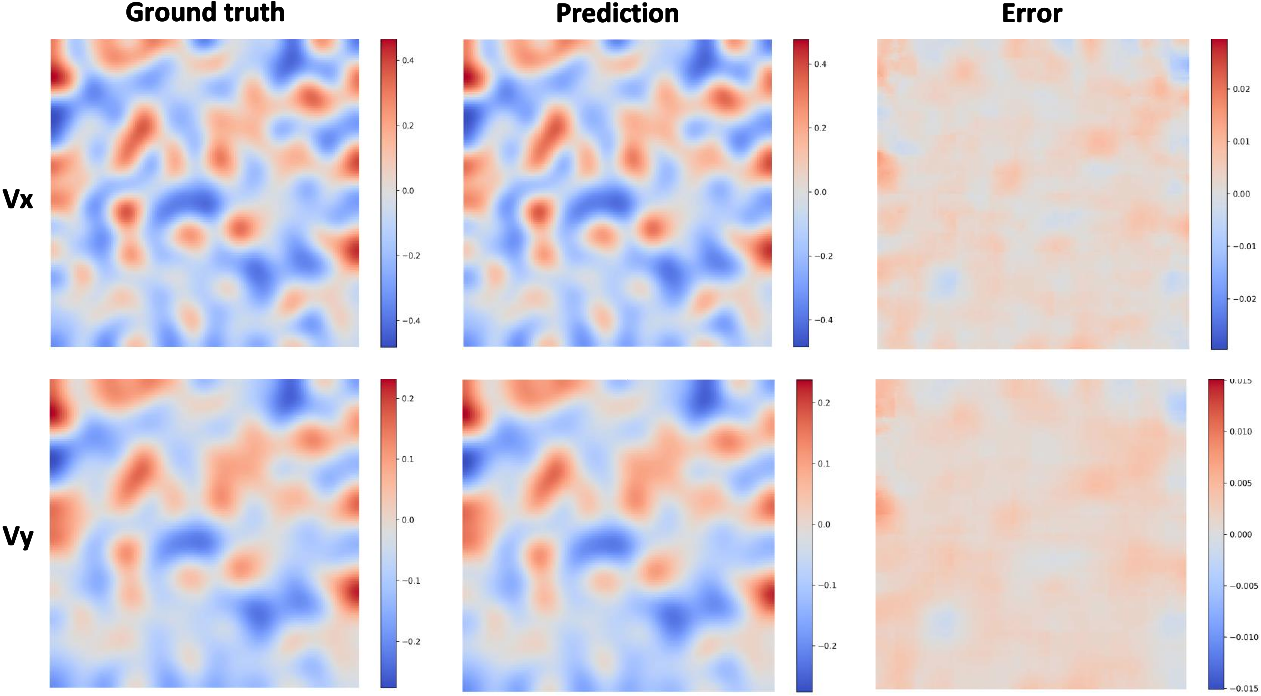}
    \caption{DR series of result visualizations. (1) The first column shows the true value, the second column shows the model prediction value, and the third column shows the corresponding error. (2) Each row is the predicted physical quantity.}
    \label{fig:dr}
    \vspace{-12pt}
\end{figure}

\begin{figure}[t]
    \centering
    \includegraphics[width=0.75\linewidth]{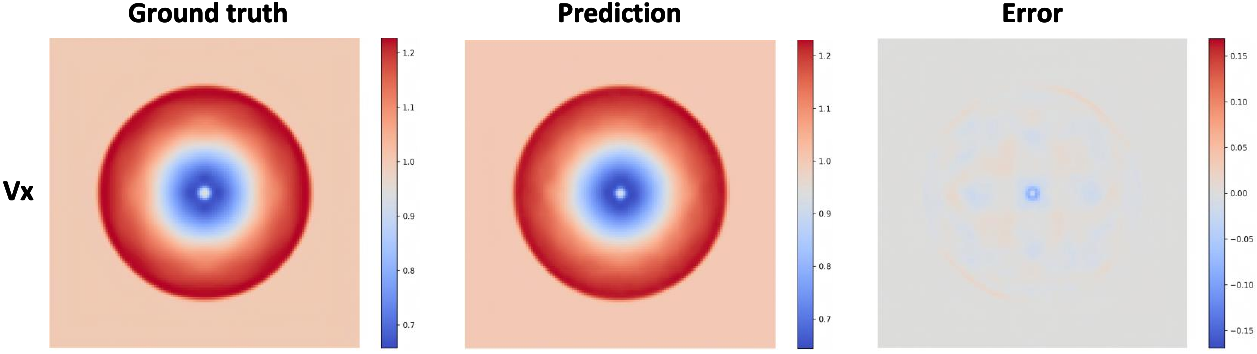}
    \caption{SWE series of result visualizations. (1) The first column shows the true value, the second column shows the model prediction value, and the third column shows the corresponding error. (2) Each row is the predicted physical quantity.}
    \label{fig:swe}
    \vspace{-12pt}
\end{figure}

\begin{figure}[t]
    \centering
    \includegraphics[width=0.75\linewidth]{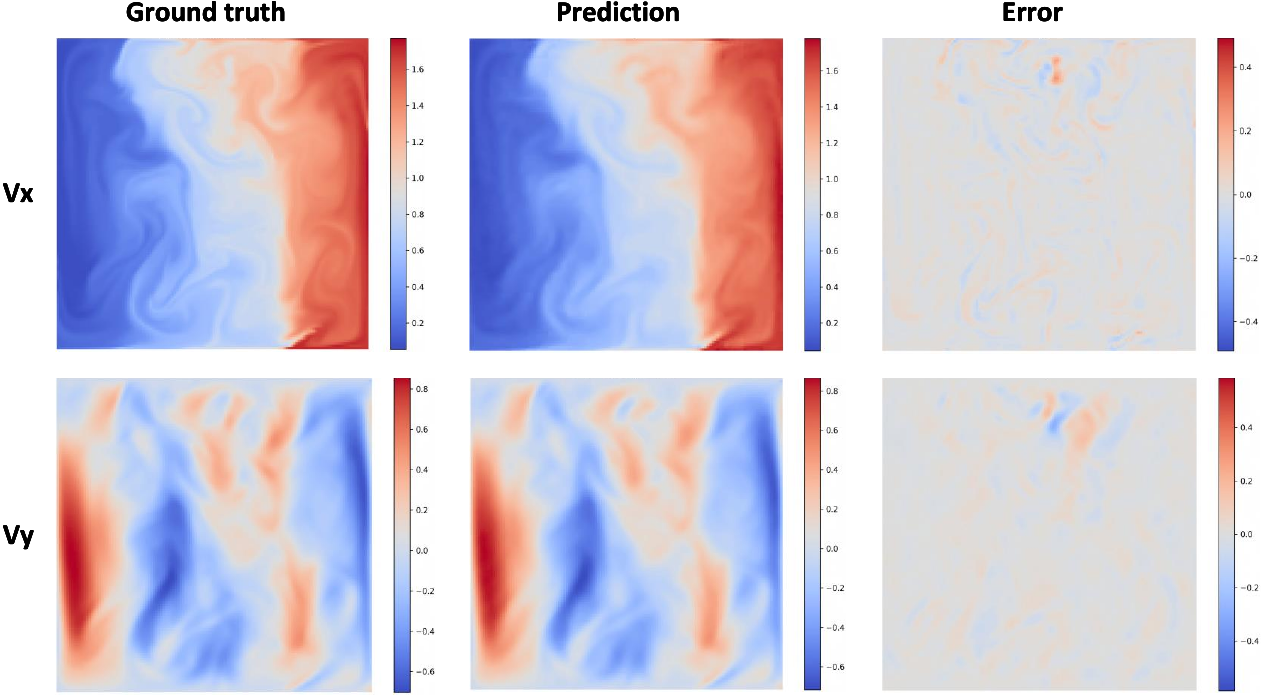}
    \caption{PDEArena series of result visualizations. (1) The first column shows the true value, the second column shows the model prediction value, and the third column shows the corresponding error. (2) Each row is the predicted physical quantity.}
    \label{fig:ns}
    \vspace{-12pt}
\end{figure}

\begin{figure}[t]
    \centering
    \includegraphics[width=0.75\linewidth]{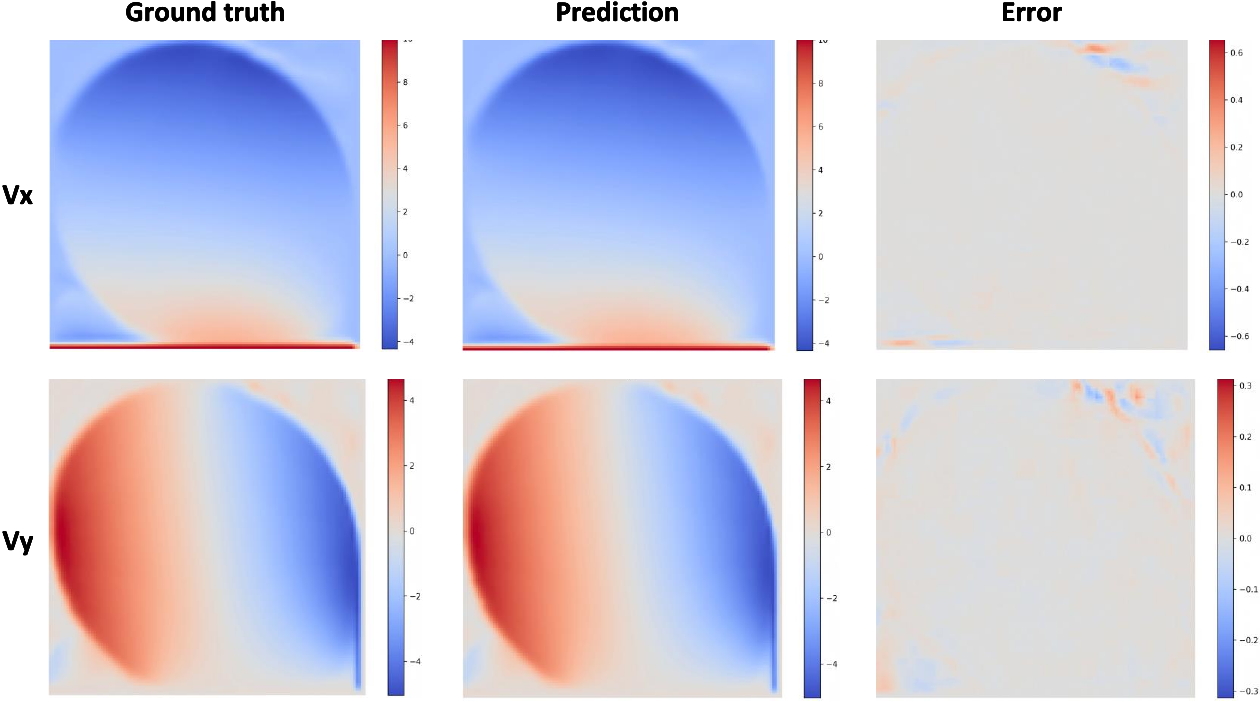}
    \caption{CFDBench series of result visualizations. (1) The first column shows the true value, the second column shows the model prediction value, and the third column shows the corresponding error. (2) Each row is the predicted physical quantity.}
    \label{fig:cfdbench}
    \vspace{-12pt}
\end{figure}


\end{document}